# DSM Refinement with Deep Encoder-Decoder Networks

NANDO METZGER[1]

*3D city models can be generated from aerial images. However, the calculated DSMs suffer from noise, artefacts, and data holes that have to be manually cleaned up in a time-consuming process. This work presents an approach that automatically refines such DSMs. The key idea is to teach a neural network the characteristics of urban area from reference data. In order to achieve this goal, a loss function consisting of an L1 norm and a feature loss is proposed. These features are constructed using a pre-trained image classification network. To learn to update the height maps, the network architecture is set up based on the concept of deep residual learning and an encoder-decoder structure. The results show that this combination is highly effective in preserving the relevant geometric structures while removing the undesired artefacts and noise.*

## 1 Motivation

3D city models are an integral part of online map services and have a wide range of applications such as navigation, urban planning, insurances, facility management, entertainment industry, virtual reality, routing, 3D cadastral systems, noise simulations, and many more. A favoured way to create a digital surface model (DSM) is to use dense photogrammetric reconstruction. Matching pixels in images and estimating and fusing depth maps are fully automated processes. However, the resulting DSMs contain artefacts as the comparison of Figure 1 shows. In the automatically generated DSM (Figure 1, left), the surfaces are noisier than in the reference DSM (Figure 1, right). Another issue is that the dense matching algorithm can have problems in resolving ambiguities in the images. The building in the lower part of Figure 1 has multiple similar-looking roof dormers. The dense matching algorithm has difficulties matching the right dormer in every aerial image which leads to systematic outliers like the holes in the building. Furthermore, to reconstruct the coordinates of an object point, the point has to be visible and detectable in at least two aerial images. Due to shadowing effects from buildings and trees, some points cannot be reconstructed which results in lack of data in some areas. Note that the area shown in Figure 1 does not include this issue.

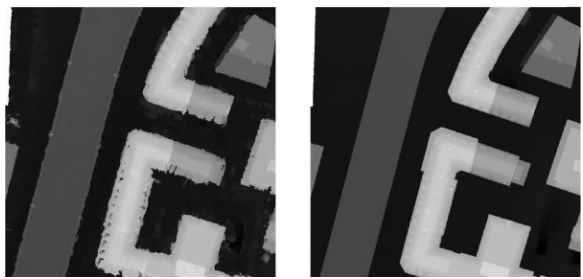

Figure 1: Raw DSM generated by automatic dense matching of aerial imagery (left) and a manually cleaned up DSM (right) displayed as height maps (brighter gray values have larger heights).

---

[1] BSC of Geomatics, ETH Zürich, Switzerland; metzgern@ethz.ch

Classical matching algorithms typically include an explicit preference for piece-wise smooth surfaces (HIRSCHMÜLLER 2005). However, such a prior is rather vague and knows very little about the geometric layout of the observed scene. In the context of urban modelling, buildings are likely to have parallel walls, dormers have characteristic dimensions, building facades are perpendicular to the street, and so on. Encoding these soft prior expectations about urban scenes into a mathematical constraint is difficult. The alternative is to learn the characteristics of urban areas directly from data. State-of-the-art image processing approaches for learning such complex patterns are based on deep learning neural networks.

The goal of this work is to enhance DSMs generated by dense photogrammetric reconstruction by learning an *a priori* model of urban structures from data. The idea is to train a deep convolutional neural network that refines an initial DSM by completing missing geometry, removing noise, and enhancing missing details. These steps are currently performed in a manual manner. The automation of these steps makes the generation of 3D city models a more efficient process.

It is important to note that the model does not have to lean basic 3D reconstruction as the input is already a DSM. The approach is to start with an initial DSM $X$ and learn to predict residuals $f(X)$ to update the DSM as shown in Figure 2.

The output $\hat{y}$ is compared to the reference data $y$ using a loss function to get a measure of discrepancy. The measured discrepancy is then fed back to the model via backpropagation in order to update its weights and to improve the prediction of the model.

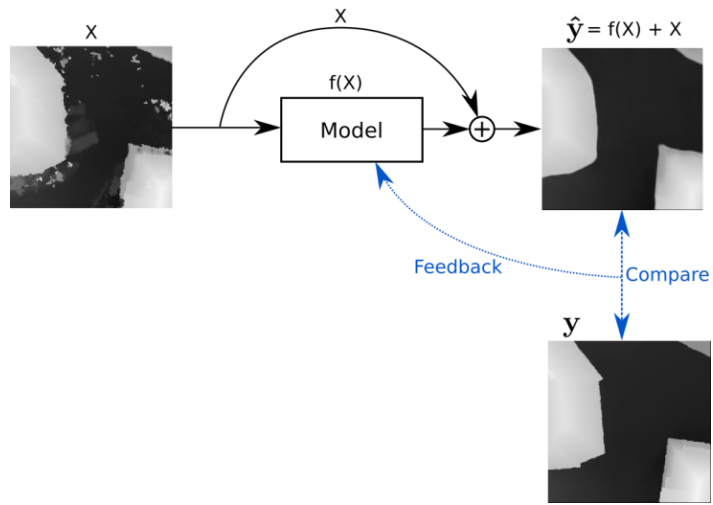

Figure 2: Proposed concept of residual learning and feedback loop.

## 2 Methodology

### 2.1 Data

The input data of the model is a photogrammetric digital surface model of the city of Zurich in Switzerland. The image data set is provided by the ISPRS/EUROSDR (2014) *Benchmark on High Density Image Matching*. The ground sampling distance of the areal images is specified as 6-13 cm and the covered area is approximately 1×1 km. The images are processed using the photogrammetric surface reconstruction software SURE (NFRAMES 2019), where a resolution of 10 cm is used for the computation of the DSM. Near the border of the project area, there are fewer overlapping images. This means that the quality of the DSM is lower at the border than in the centre of the image block because less points can be reconstructed from the image data set. When also taking into account the shadowing effects due to buildings and vegetation, holes of missing data are very likely to appear in these areas. These holes have to be interpolated in an additional preprocessing step. Furthermore, vegetation poses challenges to the dense matching and depth map estimation procedure due to the repetitive texture of tree crowns and grass. If the same tree is photographed from slightly different angles, the pattern of the leaves changes due to coverages and shadows. This creates noise and outliers in the DSM.

As ground truth data, the 2.5D city model provided by (OPEN DATA 2018) is chosen. The data set consists of a LOD2 model, which contains buildings with detailed roof structures and a terrain model. The data is gained from photogrammetric measurements and is processed in a semi-automatic manner.

To train the neural network, the whole area is cut into smaller patches. The entire dataset is split into three mutually exclusive areas, where one area is used for training, one for validation, and one for testing. The validation and testing area both have a share of 9%. The training patches are cropped from random positions of the training region in order to favor overlapping images, such that more samples can be created, whereas the testing and validation patches are cut in a grid. The size of 256×256 px, which corresponds to an area of ~ 25×25 m in world coordinates, has empirically proven to be a reasonable patch size.

In order to enlarge the training data set and robustify the behaviour of the final model, data augmentation is applied to the training patches. The augmentation includes random rotations as well as mirroring. Translation is already realized by the random sampling positions of the patches. Scaling and shearing of the images have been omitted because these operations could distort the architectural characteristics.

As a last preprocessing step, the data is normalised by centring to height 0 (per patch) and scaling by the global standard deviation of the heights derived from the training set. Data normalisation encourages more stable training and faster convergence.

## 2.2 Network Architecture

The architecture is an encoder-decoder network (RONNEBERGER et al. 2015) with skip connections at each level of depth. The network consists of 43,265,345 trainable weights. The model is fully-convolutional which means that the number of parameters is independent of the input/output patch size. In this way, the model can be trained on smaller patches, while predictions can be made with larger sized patches. The trainable parameters are the kernels and biases of the convolutional layers, and the parameters of the PReLU activations. Each black box in Figure 3 represents a set of feature maps (tensors). The number of channels is specified at the top or bottom of the box and the width and height are specified at the side of the boxes.

The left part of Figure 3 is called the encoder. It is built from a convolutional layer (blue arrows) and a MAX pooling operation (red arrows) to reduce the spatial dimension. The convolutional layers have a stride of 1, a 4×4 kernel, and operate with zero-padding. ReLU functions have the risk of becoming inactive for negative input values, hence, the PReLU function is used (HE et al., 2015). The MAX pooling operation has a 2×2 kernel and stride of 2. The idea is that the convolutional layer specialises on pattern recognition and the MAX-pooling layer reduces the spatial dimension while keeping important information.

The decoder (right half of Figure 3) consists of convolutional layers with ReLU activations (purple arrows) and upsampling operations by repetition of pixels (green arrows). Here, the inactivity of the ReLU can be beneficial, since the ReLU is able to set output values to zero.

The last operation of the network is a convolutional layer with a linear activation (orange arrow). It is important to ensure that the final output of the model is unbounded such that every real value could be predicted. The skip connections (black arrows) add the output of the convolutional layers in the encoder block to the corresponding output of the upsampling layers in the decoder. Skip connections help to preserve high frequency details that are generally lost during the downsampling process of the encoder.

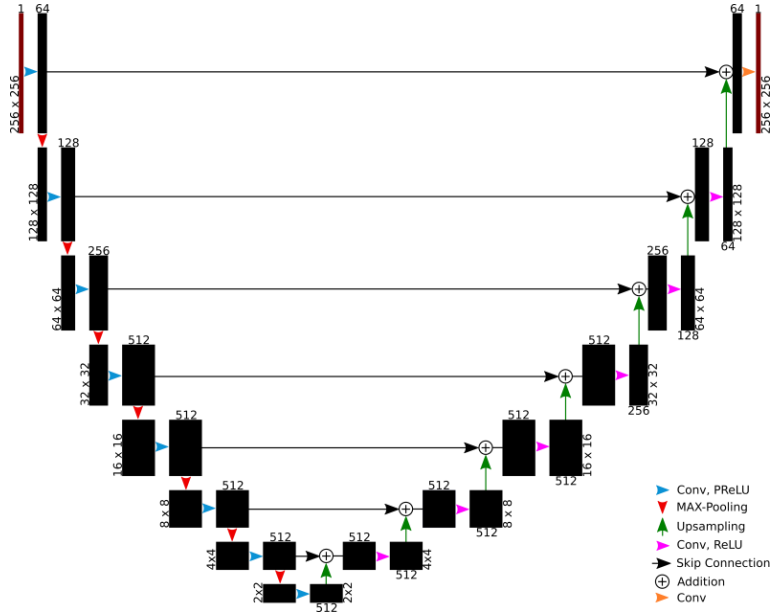

Figure 3: Proposed network architecture.

## 2.3 Loss Metric

The choice of the loss function is a crucial part of any deep learning method. The constructed loss function shown in Equation 1 consists of four terms, which are described in this section. The factors $\lambda$ are scaling factors to combine the individual terms, which have to be optimized as hyperparameters.

$$\begin{aligned}
\mathcal{L}_{total} &= \lambda_{img}\mathcal{L}_{img} + \lambda_{weights}\mathcal{L}_{weights} + \lambda_{activity}\mathcal{L}_{activity} + \lambda_{feat}\mathcal{L}_{feat} \\
&= \frac{1}{N_{img}}\lambda_{img}\|\hat{\mathbf{y}} - \mathbf{y}\|_1 + \lambda_{weights}\|\mathbf{w}\|_1 + \lambda_{activity}\|\mathbf{a}\|_1 + \lambda_{feat}\mathcal{L}_{feat}
\end{aligned} \quad (1)$$

The used data set contains edges and corners contaminated by noise of an unknown distribution. Furthermore, outliers due to ambiguities in the dense matching algorithm exist. Applying the often used L2 loss function assumes that the noise is Gaussian distributed. Furthermore, the L2 loss function tends to favour solutions that smooth out the values. This effect is highly undesirable for the task of DSM refinement because it leads to smoothed building edges, and thus, the characteristic shape of the buildings cannot be preserved. Therefore, for this task, the L1 loss function is used, since it has no such bias towards smoothness. This pixel-wise loss is denoted as $\mathcal{L}_{img}$ in Equation 1. $\hat{y}$ and $y$ denote the prediction and the ground truth respectively. $N_{img}$ is the number of pixels and $w$ are the weights of the kernels.

In addition to the pixel-wise L1 loss, a penalty regularisation is used as shown in Equation 1. The L1 weight regularisation (denoted as $\mathcal{L}_{weights}$), encourages sparse weights (FRIEDMANN et al. 2008). The L1 activity regularisation (denoted as $\mathcal{L}_{activity}$) is applied to the output values of the decoder (denoted as $a$) only. Recall that the network is a residual network. Hence, the term $\mathcal{L}_{activity}$ encourages the network to produce sparse residuals.

This loss function is extended with a perceptual loss term (denoted as $\mathcal{L}_{feat}$). The perceptual metric is obtained by comparing features from the pertrained VGG16 network (SIMONYAN AND ZISSERMAN 2014). To calculate the perceptual loss term, both the prediction of the encoder-decoder network and the ground truth patch are fed into the VGG16 network. The calculated features of five intermediate layers are picked and compared to each other. The chosen layers are distributed over the whole network to figure out which of the VGG16 layers are the most suitable for constructing the perceptual loss. The features are compared via a L1 norm and scaled by the number of features per layer $N_i$ as shown in Equation 2. The symbol $\mathcal{F}$ denotes the set of layers of the VGG16 network that are selected. $\hat{y}_{feat,i}$ and $y_{feat,i}$ denote the predicted and the ground truth features respectively and $\lambda_{feat,i}$ are the hyperparameters which scale the individual losses.

$$\mathcal{L}_{feat} = \sum_{i \in \mathcal{F}} \lambda_{feat,i} \mathcal{L}_{feat,i} = \sum_{i \in \mathcal{F}} \frac{1}{N_i} \lambda_{feat,i} \|\hat{y}_{feat,i} - y_{feat,i}\|_1 \tag{2}$$

## 3 Results and Discussion

Figure 4 shows the prediction performance of the final model. The model has learned that buildings are built according to simple geometric standards such as straight walls with piece-wise planar roof structures. Furthermore, the model removes vegetation and outliers from DSMs.

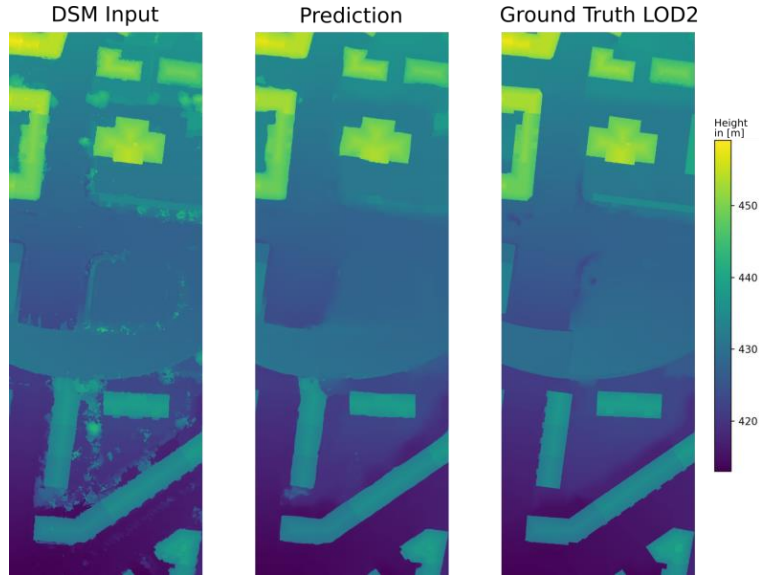

Figure 4: Prediction performance on the test area. From left to right: given input, prediction of the model, corresponding ground truth patch.

However, the predictions on the test area show that the model still has some limitations. It would be desirable that smaller structures like balconies are learned too. Furthermore, the model still predicts walls that are not exactly straight and vegetation is not always removed completely. The model achieves to predict 74% of the test pixels correctly. A pixel is considered as correct if the difference between the predicted and ground truth height is less than 0.50 cm. The mean absolute error of the refined DSM is 0.65 m, while the median absolute error is 0.23 m.

## 4  Conclusion

This work addresses the problem of urban DSM refinement using a learning-based approach. It was possible to implement a neural network, which cleans the artefacts of the photogrammetric generated DSM. The key idea is to learn typical urban structures from reference data instead of using hard-crafted priors. This paper proposes an encoder-decoder architecture based on the concept of residual learning as a method to update noisy input DSMs. A suitable loss function, which contains a perceptual metric, is constructed in order to train the network.

The trained model can be applied to unseen data sets with only a few restrictions. Additional training data from other regions is required to adapt to the local style of urban planning and architecture. Furthermore, the model is trained with a pixel size of 10 cm. The performance will decrease if the model is applied to a DSM whose GSD differs from 10 cm.

Feeding in the raw aerial images into the model could improve the performance. Furthermore, a larger training area would give the network more opportunities to learn urban structures.